\documentclass[letterpaper, 10 pt, conference]{ieeeconf}  

\IEEEoverridecommandlockouts                              

\IEEEoverridecommandlockouts                              

\usepackage[english]{babel}
\usepackage{amsmath}
\usepackage{amssymb}
\usepackage{booktabs}
\usepackage{siunitx}
\usepackage[dvips]{graphicx}
\usepackage{multirow}
\usepackage{booktabs}
\usepackage{makecell}
\usepackage{siunitx}
\usepackage{amsfonts}
\usepackage{enumerate}
\usepackage{tabularx}
\usepackage{algorithm,algorithmic}
\usepackage{bm}
\usepackage{adjustbox}
\usepackage{xcolor}
\usepackage{siunitx}
\usepackage{booktabs}
\usepackage{mdframed}
\usepackage{adjustbox}
\usepackage{authblk}
\usepackage{color}
\usepackage{xurl}
\usepackage{subcaption}
\usepackage{cite}
\makeatletter
\let\NAT@parse\undefined
\makeatother
\usepackage{hyperref}
\usepackage{relsize}
\usepackage{float}
\usepackage{pifont}
\usepackage{bm}
\newcommand{\reffig}[1]{Fig.~\ref{#1}}

\newcommand{\refequ}[1]{Eq.~\eqref{#1}}

\usepackage{acro}
\DeclareAcronym{PMP}{
  short = PMP,
  long  = Pontryagin's Maximum Principle,
  short-indefinite = a,
  long-indefinite = a
}

\DeclareAcronym{DEM}{
  short = DEM,
  long  = Digital Elevation Map
}

\DeclareAcronym{FMU}{
  short = FMU,
  long  = Flight Management Unit
}

\DeclareAcronym{OMPL}{
  short = OMPL,
  long  = Open Motion Planning Library
}

\DeclareAcronym{RMSE}{
  short = RMSE,
  long  = Root Mean Squared Error
}

\DeclareAcronym{ROS}{
  short = ROS,
  long  = Robot Operating System
}

\DeclareAcronym{VTOL}{
  short = VTOL,
  long  = Vertical Take-Off and Landing,
  short-indefinite = a,
  long-indefinite = a
}
\usepackage{subcaption}

\title{\LARGE \bf
Minimum Time Dubins Airplane Paths with Asymmetric Climb Rates
}

\author{Jaeyoung Lim and Giuseppe Loianno
\thanks{The authors are with Department of Electrical Engineering and Computer Sciences, University of California, Berkeley, CA 94720, USA.
        {\tt\footnotesize email: \{jaeyounglim, loiannog\}@berkeley.edu}.}%
\thanks{This work was supported by the DARPA Albatross Grant HR00112590173. Approved for Public Release, Distribution Unlimited. The views, opinions and/or findings expressed are those of the authors and should not be interpreted as representing the official views or policies of DARPA or the U.S. Government.}
}

\begin{document}

\maketitle
\thispagestyle{empty}
\pagestyle{empty}

\begin{abstract}
Dubins airplane paths approximate the limited maneuverability of fixed-wing vehicles with minimum curvature and climb rate constraints. However, the symmetric climb rate constraints result in sub-optimal paths and conservative vehicle performance. In this work, we propose \emph{asymmetric Dubins airplane paths}, which consider asymmetric climb rates for climbing and descending. We revisit the time optimality conditions and show that the asymmetric flight path angle constraints preserve optimality. We show that by considering asymmetric climb rates, we can take advantage of full performance of the vehicle, reducing the minimum time by 71\% for connecting randomly generated states. We also demonstrate that the added climb rate results in 2.8$\pmb{\times}$ faster to find the median solution time when integrated into a sampling-based planning task on rugged terrain, due to the added feasibility. We further demonstrate the practicality of the approach with a real-world flight.
\end{abstract}

\section{Introduction}
Fixed-wing and hybrid \ac{VTOL} aerial vehicles play a critical role in large-scale environment monitoring~\cite{lim2026autonomous} and search and rescue~\cite{oettershagen_robotic_2018}. This is due to the energy efficiency of using aerodynamic forces to stay airborne, making them suitable for tasks requiring long endurance and long area coverage. However, limited control authority and high operating speeds constrain their maneuverability, imposing kinematic constraints such as minimum turning radius and climb rate limits~\cite{chitsaz2007time}.

Minimum turning radius is a common constraint in a variety of non-holonomic systems, such as cars~\cite{reeds1990optimal}, underwater vehicles~\cite{moll20243d}, and fixed-wing vehicles~\cite{chitsaz2007time}. Dubins curves~\cite{dubins1961plane} have been widely used by providing closed-form solutions for curvature-constrained shortest paths for planning~\cite{moll20243d, phillips2021learn}. Dubins airplane paths~\cite{chitsaz2007time, mclain_implementing_2015} extend the Dubins curves into three-dimensional space, by taking into account symmetric climb rate constraints. Dubins airplane paths have been successfully deployed in real world fixed-wing navigation scenarios in confined environments~\cite{lim2026autonomous, lim2024safe} as they are efficient to compute. However, fixed-wing vehicles typically have asymmetric climb rate constraints, with different limits when climbing versus descending. 
This is especially true for aerodynamically efficient high-aspect-ratio platforms, which can typically climb at much steeper angles than they can descend.
Therefore, practical implementations~\cite{lim2024safe} typically use a conservative lower value of the climb rate constraints to satisfy the symmetric climb rate constraint assumed by the Dubins airplane model. This results in reduced feasibility and longer less efficient paths. 

In this work, we propose an \emph{asymmetric Dubins airplane paths} approach, which extends the Dubins airplane model to support asymmetric climb rate constraints.  
We revisit the optimality analysis presented in~\cite{chitsaz2007time}, and validate that the extremal solutions under asymmetric climb rate constraints preserve optimality. Subsequently, we evaluate performance on downstream tasks, such as regulation-compliant steep-terrain navigation, and show that the added feasibility from the asymmetric climb rates results in faster solve times and more shorter paths. We further demonstrate our method through real-world outdoor flight experiments and show that the vehicle can successfully follow the path. 

\begin{figure}[t]
    \centering
    \begin{subfigure}{0.48\linewidth}
    \centerline{\includegraphics[width=\linewidth]{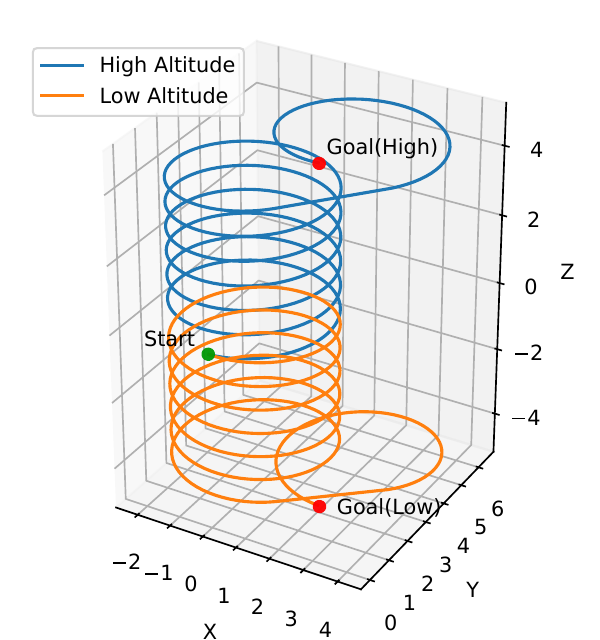}}
    \caption{Symmetric climb rates}
    \label{fig:test_vehicle}
    \end{subfigure}
    \begin{subfigure}{0.48\linewidth}
    \includegraphics[width=\linewidth]{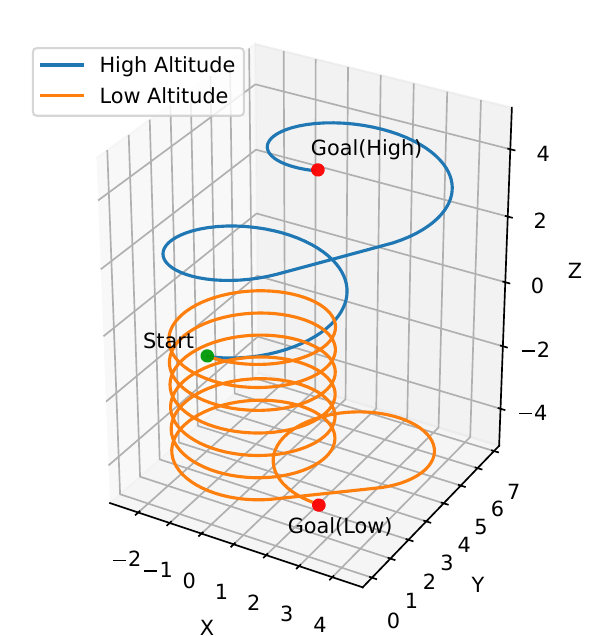}
    \caption{Asymmetric climb rates}
    \label{fig:test_vehicle}
    \end{subfigure}
    \caption{Dubins airplane paths with a) symmetric climbrates~\cite{chitsaz2007time} and b) asymmetric climbrates (proposed) between start (green) and goal (red). It can be seen that the asymmetric climb rates allow steeper flight path angles, reducing the time required to reach the target configuration.}
    \label{fig:block_diagram}
\end{figure}

The contributions of this paper are as follows
\begin{itemize}
    \item Minimum time Dubins airplane paths with asymmetric climb rate constraints, exploiting the full climb and descent performance of fixed-wing vehicles. 
    \item Validation that asymmetric climb rate bounds preserve optimality, with performance gains bounded by the vehicle's climb rate asymmetry ratio.
    \item Demonstration of path representation on a rugged terrain planning task, showing $2.8\times$ faster median planning time and real-world flight demonstration.
\end{itemize}

\section{Related Works}
Dubins~\cite{dubins1961plane} showed that a curvature constrained shortest path in $SE(2)$ consists of a sequence of straight lines and arcs with maximum curvature. 
\emph{Dubins airplane paths}~\cite{chitsaz2007time} extend the Dubins curves to three-dimensional space by augmenting the state space with altitude and considering an additional climb rate constraint. The difference in altitude is classified into three cases, depending on the length of the horizontally projected Dubins curve. However, the optimality analysis is valid under the assumption that climb rates are symmetric for climbing and descending. Therefore, practical deployments conservatively adopt the lower of the two climb-rate bounds~\cite{lim2024safe, duan2024energy}. 
Asymmetric flight path angles have been considered, but at the cost of optimality~\cite{mclain_implementing_2015} or large lateral detours from employing an additional Dubins path along the longitudinal axis~\cite{hague2023planning, vavna2020minimal}. Other works have considered alternative continuous path representations such as polynomials~\cite{bry2015aggressive, morando2025trajectory}, but do not account for climb-rate performance limits. 
While these path representations may be useful, the closed form solution of Dubins curves makes \emph{Dubins airplane paths} efficient to compute, which is beneficial for its use in sampling-based planners. In this work, we focus on extending the \emph{Dubins airplane path} representation to accommodate asymmetric climb rate constraints to be able to exploit the full performance of fixed-wing aircraft. 

\section{Problem Formulation}
\subsection{Minimum Time Dubins Airplane Paths}
Let us consider a vehicle state $\bm{x} = (x, y, z, \theta) \in \mathbb{R}^3 \times SO(2)$, where $x, y, z$ is the position and $\theta$ is the heading. Given that the vehicle is controlled via two bounded control inputs $\bm{u} = (u_z, u_{\theta})$, where $u_z$ is the climb rate and $u_{\theta}$ heading rate subject to bounds $|u_{\theta}| < u_{\theta max}$, $u_z \in |u_z| < u_{zmax}$, the kinematics of the Dubins airplane model can be written as \refequ{eq:dubins_airplane_model}.
\begin{align}
    \Dot{\bm{x}} = \begin{pmatrix}\Dot{x}\\\Dot{y}\\\Dot{z}\\ \Dot{\theta}\end{pmatrix} = f(\bm{x}, \bm{u}) = 
        \begin{pmatrix}V\cos(\theta) \\
         V\sin(\theta) \\
         u_{z} \\
        u_{\theta}  \end{pmatrix}.
        \label{eq:dubins_airplane_model}
\end{align}
\begin{figure}[b]
    \centering
    \includegraphics[width=0.6\linewidth]{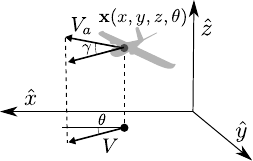}
    \caption{State space representation of a Dubins airplane model.}
    \label{fig:dubins_airplane}
\end{figure}

It has been shown that the minimum time path between two states, $\bm{x}_i=(x_i, y_i, z_i)$, $\bm{x}_t=(x_t, y_t, z_t)$ can be found by classifying the altitude difference into three cases: \emph{low altitude}, \emph{medium altitude}, and \emph{high altitude}~\cite{chitsaz2007time}. For the \emph{low Altitude} case, the required altitude change can be achieved within the climb-rate constraint while traversing the 2D Dubins path. The minimum time is therefore given by the time to traverse the 2D Dubins distance $D(\bm{x}_i, \bm{x}_t)$ at ground speed $V$~(\refequ{eq:low_altitude}).
\begin{align}
    \Delta T_{Low} = D(\bm{x}_i, \bm{x}_t)/V.
    \label{eq:low_altitude}
\end{align}
For the \emph{medium altitude} and \emph{high altitude} cases, the minimum time is instead determined by the time required to achieve the altitude difference at the maximum climb rate. While all paths that match the path length will be optimal paths, we follow the approach in~\cite{mclain_implementing_2015}, where the 2D Dubins path is appended with a helix or an additional arc depending on whether the altitude difference falls into \emph{high altitude} or \emph{medium altitude} case.
\begin{align}
    \Delta T_{High, Medium} = |z_i - z_t| / u_{zmax}.
\end{align}
Note that the climb rate constraint $u_{zmax}$ is symmetric whether the vehicle needs to climb ($z_t > z_i$) or descend ($z_t < z_i$).

\begin{figure}[t]
    \centering
    \includegraphics[width=0.9\linewidth]{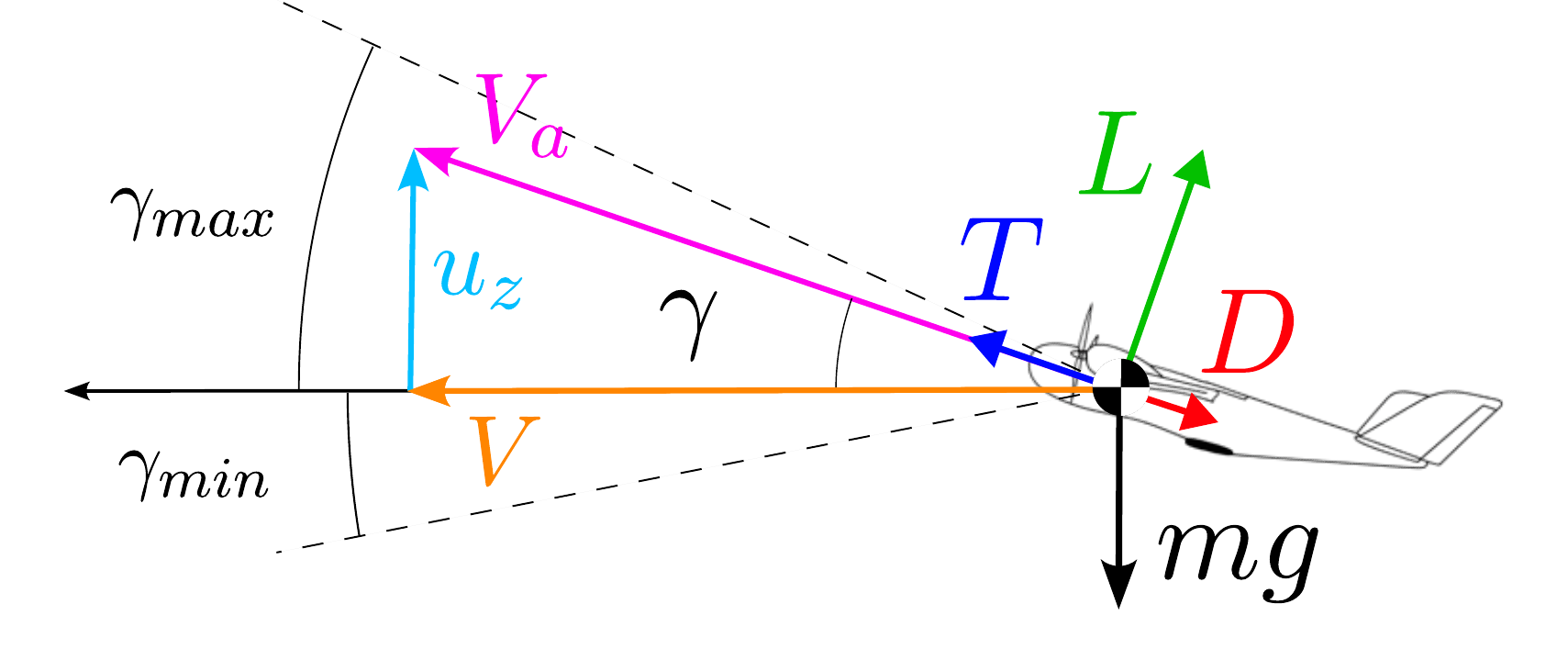}
    \caption{Force diagram of static equilibrium of the fixed-wing aircraft model. Groundspeed $V$ is assumed to be constant, and the flight path angle $\gamma$ is determined by the climbrate $u_z$, which is bounded by the maximum climbrate $\gamma_{max}, \gamma_{min}$. }
    \label{fig:force_balance}
\end{figure}

\subsection{Performance Limits of an Aircraft}
We show how the symmetric climb rate constraint is not well aligned with performance limit of fixed-wing vehicles. To determine the performance limits of a fixed-wing vehicle, we consider the vehicle in a static equilibrium~(\reffig{fig:force_balance}). In a static equilibrium, the thrust $T$, lift $L$, drag $D$, and weight $mg$ are in balance as expressed by
\begin{align}
    T - D - mg \sin{\gamma} &= 0\label{eq:force_balance},\\
    L - mg \cos{\gamma} &= 0\nonumber.
\end{align}
Given a flight path angle $\gamma$, and airspeed $V_a$, the climb rate $u_z$ of the vehicle can be calculated as
\begin{align}
    u_z = V_{a} sin \gamma = V_{a} \frac{T - D}{mg},\label{eq:rate_of_climb}
\end{align}
where $V_a(T-D)/mg$ is the specific power, which is the power injected into the system normalized by the weight of the vehicle. Note that for the Dubins airplane model, we assume the horizontal speed $V$ to be constant, instead of the airspeed $V_a$. While \cite{mclain_implementing_2015} considers the flight path angle limit with constant $V_a$, we keep the groundspeed constant to preserve the structure of the hamiltonian~\cite{chitsaz2007time}. The lift and drag can be determined by calculating the airspeed for a specific flight path angle $V_a =  V/\cos{\gamma}$.

Since the thrust is controllable within $T \in [0, T_{max}]$, the maximum climb rate $u^+_{z_{max}}$ and descend rate $u^-_{z_{max}}$ can be computed as~\refequ{eq:min_max_climbrate}.
\begin{align}
    u^+_{z_{max}} &=V_a\sin{\gamma_{max}} =  V_a\frac{T_{max} - D}{mg},\\
    u^-_{z_{max}} &=V_a\sin{\gamma_{min}} = -V_a\frac{D}{mg}.
    \label{eq:min_max_climbrate}
\end{align}
Maximum climb rate is defined by specific excess power $V_a(T_{max} - D)/mg$, while the maximum descend rate is defined by the specific drag $D/mg$ of the vehicle. Therefore, the climb rate is highly asymmetric depending on how much thrust is available and how aerodynamically efficient the vehicle is, as characterized by its drag.


\section{Optimal Dubins Airplane Model}
We consider the \ac{PMP} to obtain the optimality of the Dubins airplane model used in~\cite{chitsaz2007time}. The system equation in~eq.~\ref{eq:dubins_airplane_model} can be written as 
\begin{align}
    \Dot{\bm{x}} = f(\bm{x}, \bm{u}) = f_0(\bm{x}) + u_z f_1(\bm{x}) + u_{\theta} f_2(\bm{x})\label{eq:systems_dubins_airplane}\\
    f_0 = \begin{pmatrix}
        V\cos{\theta}\\
        V\sin{\theta}\\
        0\\
        0
    \end{pmatrix}, f_1 = \begin{pmatrix}
        0\\
        0\\
        1\\
        0\end{pmatrix}, f_2 = \begin{pmatrix}
        0\\
        0\\
        0\\
        1\end{pmatrix} \nonumber,
\end{align}
where $|u_{\theta}| < u_{\theta_{max}}$, $|u_{z}| \in [u^-_{z_{max}}, u^+_{z_{max}}]$.
Given the hamiltonian of the system $H(\bm{\lambda}, \bm{x}, \bm{u})$, the opimality conditions according to the \ac{PMP}~\cite{boltyanskiy1961theory}
\begin{align}
    \Dot{\lambda} &= - \frac{\partial H}{\partial x},\label{eq:hamiltonian_diff}\\
    H(\bm{\lambda}, \bm{x}, \bm{u}) &= \max_{\tau \in U} H(\lambda(t), x(t), \tau)\label{eq:max_hamiltonian}\\
    H(\bm{\lambda}, \bm{x}, \bm{u}) &= \lambda_0\nonumber.
\end{align}
Equation!\ref{eq:hamiltonian_diff} can be found for $\lambda$ as in~\cite{chitsaz2007time}
\begin{align}
    \Dot{\lambda} = \begin{pmatrix}
        c_1\\
        c_2\\
        c_3\\
        c_1 y - c_2 x + c_4
    \end{pmatrix},
\end{align}
where $c_1$, $c_2$, $c_3$, and $c_4$ are constants.
With the condition expressed by~eq.~(\ref{eq:max_hamiltonian}), we can synthesize a controller along the extremal
\begin{align}
\begin{cases}
u_z = u^+_{z_{max}} & \text{if } c_3 > 0 \\
 u_z = u^-_{z_{max}} & \text{if } c_3 < 0\\
u_z \in [u^-_{z_{max}}, u^+_{z_{max}}] & \text{otherwise}
\end{cases}.\label{eq:optimal_controller}
\end{align}
We can see that with the asymmetric climbrate constraints, the Hamiltonian is still along the extremal preserving the optimality. 
For more detailed proof in general for the Dubins airplane can be found in~\cite{chitsaz2007time}.

\section{Asymmetric Dubins Airplane Paths}
Given the minimum time controller synthesized using the \ac{PMP}, we propose \emph{asymmetric Dubins airplane paths}. We classify the altitude into \emph{low altitude}, \emph{medium altitude}, and \emph{high altitude} cases as done in~\cite{chitsaz2007time}. Given the vehicle state $\bm{x} = (x, y, z, \theta) \in \mathbb{R}^3 \times SO(2)$ that is controlled via two bounded control inputs $\bm{u} = (u_z, u_{\theta})$ we consider the problem of finding a minimum time path between states $\bm{x}_t$ and $\bm{x}_i$, where $|u_{\theta}| < u_{\theta_{max}}$ $u_z \in [u^-_{z_{max}}, u^+_{z_{max}}]$. We assume that the climb rate constraints are admissible, where $u^+_{z_{max}} > 0$ and $u^-_{z_{max}} < 0$.

\subsection{Low Altitude Case}
The \emph{low altitude} occurs when the altitude gain or loss between the initial and target configurations can be achieved within the climb rate bounds. 
The condition for the \emph{low altitude} case can be written as
\begin{align}
    u^-_{z_{max}} < \frac{z_t - z_i}{D(\bm{x}_i, \bm{x}_t)/V} < u^+_{z_{max}}.
    \label{eq:condition_low_alt}
\end{align}
This occurs when $c_3=0$ in~eq.~(\ref{eq:optimal_controller}). Therefore, the minimum path time and optimal climb rate for the low altitude case can be defined as
\begin{align}
    \Delta T^* &= \frac{D(\bm{x}_i, \bm{x}_t)}{V},\\
    u_z^* &= \frac{z_t - z_i}{\Delta T^*}\nonumber.
\end{align}
where $u_z^*$ is the constant climb rate required to achieve the altitude change $z_t - z_i$ over the 2D travel time $\Delta T^*$. 

\begin{figure}[t]
    \centering
    \includegraphics[width=0.9\linewidth]{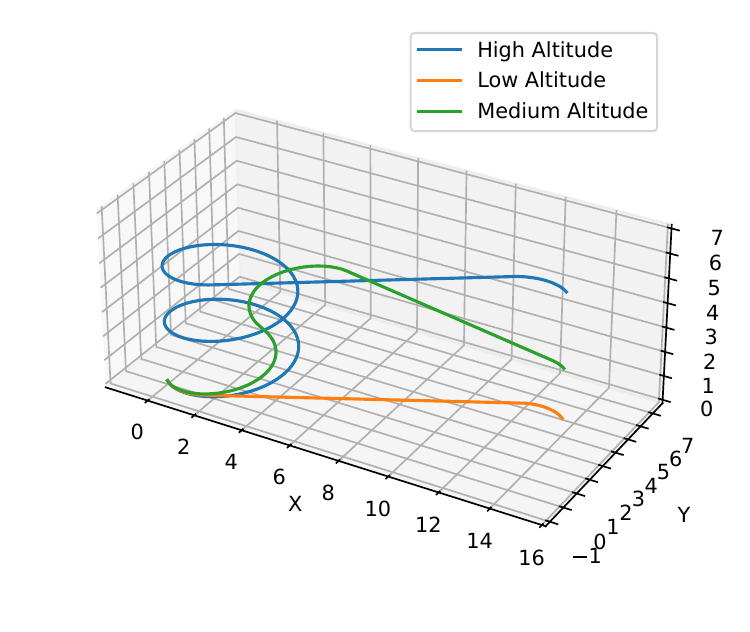}
    \vspace*{-0.7cm}
    \caption{Example of Dubins airplane paths with different altitude cases.}
    \label{fig:block_diagram}
\end{figure}

\subsection{High Altitude Case}
The \emph{high altitude} case occurs when the altitude gain exceeds what is achievable at the maximum climb rate $u^+_{z_{mmax}}$, or the required altitude loss exceeds what is achievable at the maximum descent rate $u^-_[z_{max}]$. The high altitude condition can be expressed as
\begin{align}
    u^+_{z_{max}} &< \frac{z_t - z_i}{(D(\bm{x}_i, \bm{x}_t) + 2 \pi R_{min} )/V},\\
    u^-_{z_{max}} &> \frac{z_t - z_i}{(D(\bm{x}_i, \bm{x}_t) + 2 \pi R_{min} )/V}.\nonumber
\end{align}
This corresponds to the case, where $c_3 \neq 0$ in eq.~(\ref{eq:optimal_controller}). Therefore, the climb rate is at the extremal, and the minimum time duration can be found as
\begin{align}
    \Delta T^* &= \frac{|z_t - z_i|}{|u^*_z|},\\
    u_z^* &= 
    \begin{cases}
    u^+_{z_{max}} & \text{if } z_t - z_i > 0\\
    u^-_{z_{max}} & \text{otherwise}
    \end{cases}
\end{align}
where the term $2\pi R_{min}/V$ accounts for the additional time gained by inserting a full-circle loop of minimum radius $R_{min}$.

\subsection{Medium Altitude Case}
The medium altitude case corresponds to all the configurations that are neither the \emph{high altitude} nor the \emph{low altitude} case. 
This can be written as: if $z_t - z_i > 0$,
\begin{align}
   \frac{z_t - z_i}{D(\bm{x}_i, \bm{x}_t)/V} < u^{+}_{z_{max}} < \frac{z_t - z_i}{(D(\bm{x}_i, \bm{x}_t) + 2 \pi R_{min} )/V}\nonumber.
\end{align}
or if $z_t - z_i < 0$, 
\begin{align}
   \frac{z_t - z_i}{D(\bm{x}_i, \bm{x}_t)/V} < u^{-}_{z_{max}} < \frac{z_t - z_i}{(D(\bm{x}_i, \bm{x}_t) + 2 \pi R_{min} )/V}\nonumber.
\end{align}

Since the medium altitude case also falls into $c_3 \neq 0$ in eq.~(\ref{eq:optimal_controller}), the optimal climb rate is at the extremal. However, given that the duration for reaching the target altitude is not long enough to include a full circle, an additional arc is considered and found by performing a line search~\cite{mclain_implementing_2015}. The minimum time duration can be found as
\begin{align}
    \Delta T^* &= \frac{|z_t - z_i|}{|u^*_z|},\\
    u_z^* &= 
    \begin{cases}
    u^+_{z_{max}} & \text{if } z_t - z_i > 0\\
    u^-_{z_{max}} & \text{otherwise}
    \end{cases}.
\end{align}

\section{Evaluation}
We evaluate the utility of the asymmetric Dubins airplane path by a) demonstrating the reduction time compared to the conservative climb rate constraints, and b) applying it to a path planning problem and showing that it can discover feasible paths faster with the added feasibility.

\subsection{Minimum Time Paths}
\begin{figure}[b]
    \centering
    \includegraphics[width=\linewidth]{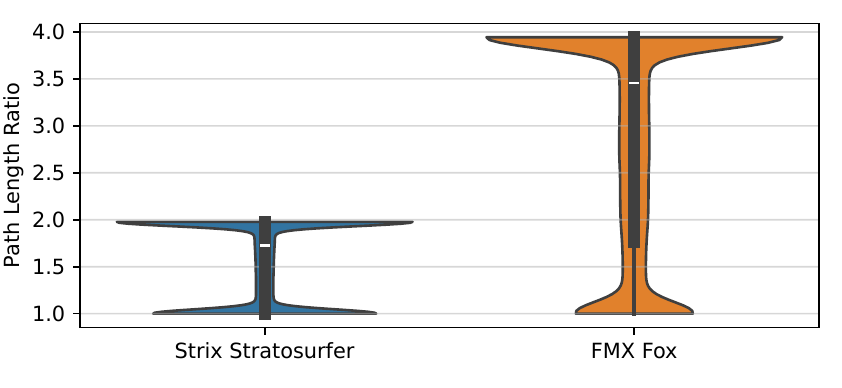}
    \caption{Distribution of the ratio of path duration between symmetric and asymmetric climb rates for example vehicle models \emph{Strix Stratosurfer}, and \emph{FMX Fox}. }
    \label{fig:benchmark_path_length}
\end{figure}

\begin{figure}[t]
    \centering
    \includegraphics[width=\linewidth]{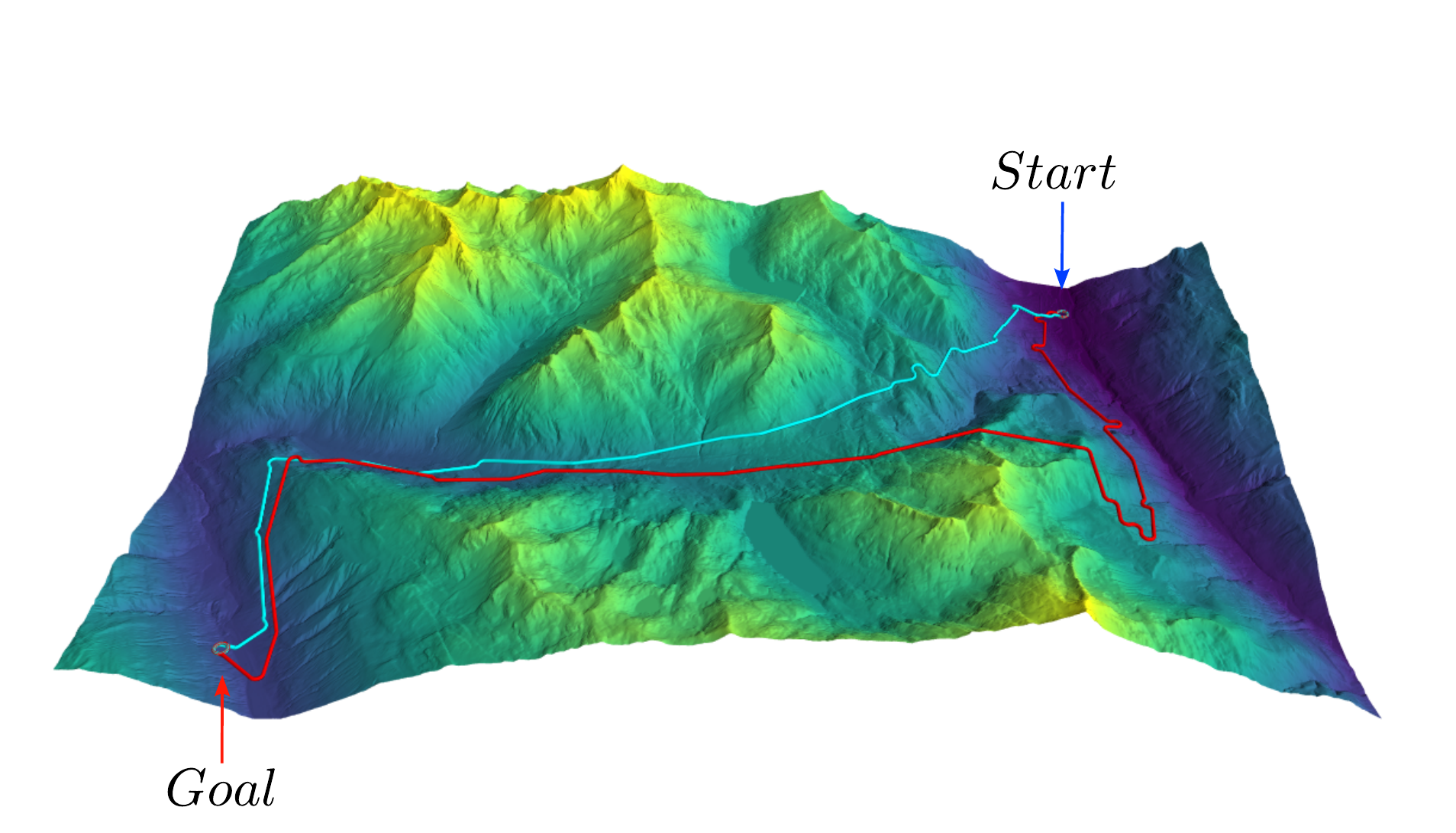}
    \caption{Qualitative comparison of the solution path from RRT* for navigating through rugged terrain using Dubins airplane paths with asymmetric climb rates (cyan) and symmetric climb rates (red) after \SI{500}{\second} of solve time. It can be seen that the path with asymmetric climb rates can follow a more aggressive climb along the terrain with the increased climb rate limit. }
    \label{fig:gotthard_pass}
\end{figure}
\begin{figure}[t]
    \centering
    \includegraphics[width=\linewidth]{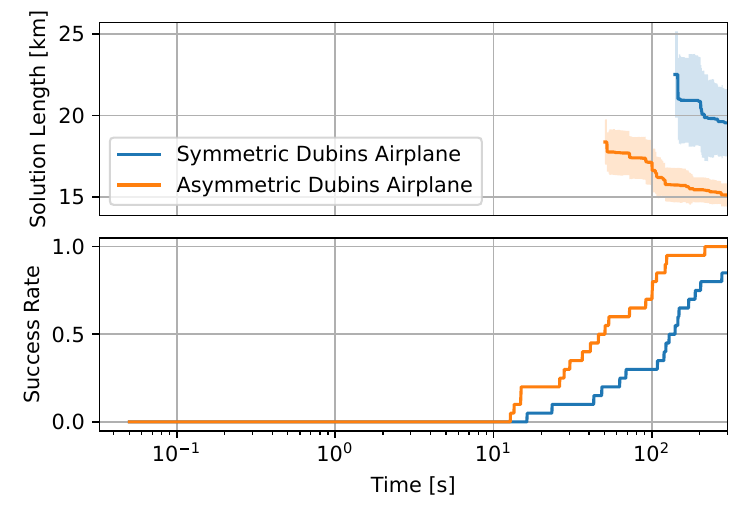}
    \caption{Comparison of median solution length and success rate compared to solve time of Dubins airplane RRT*. Median solution length is reported after \qty{50}{\percent} of solutions have been found. The added feasibility of asymmetric climb rates result in finding initial solutions faster, and better convergence in solution length.}
    \label{fig:solution_convergence}
\end{figure}
\begin{figure*}[t]
\centering
\begin{subfigure}{0.43\linewidth}
    \centering
    \includegraphics[width=\linewidth]{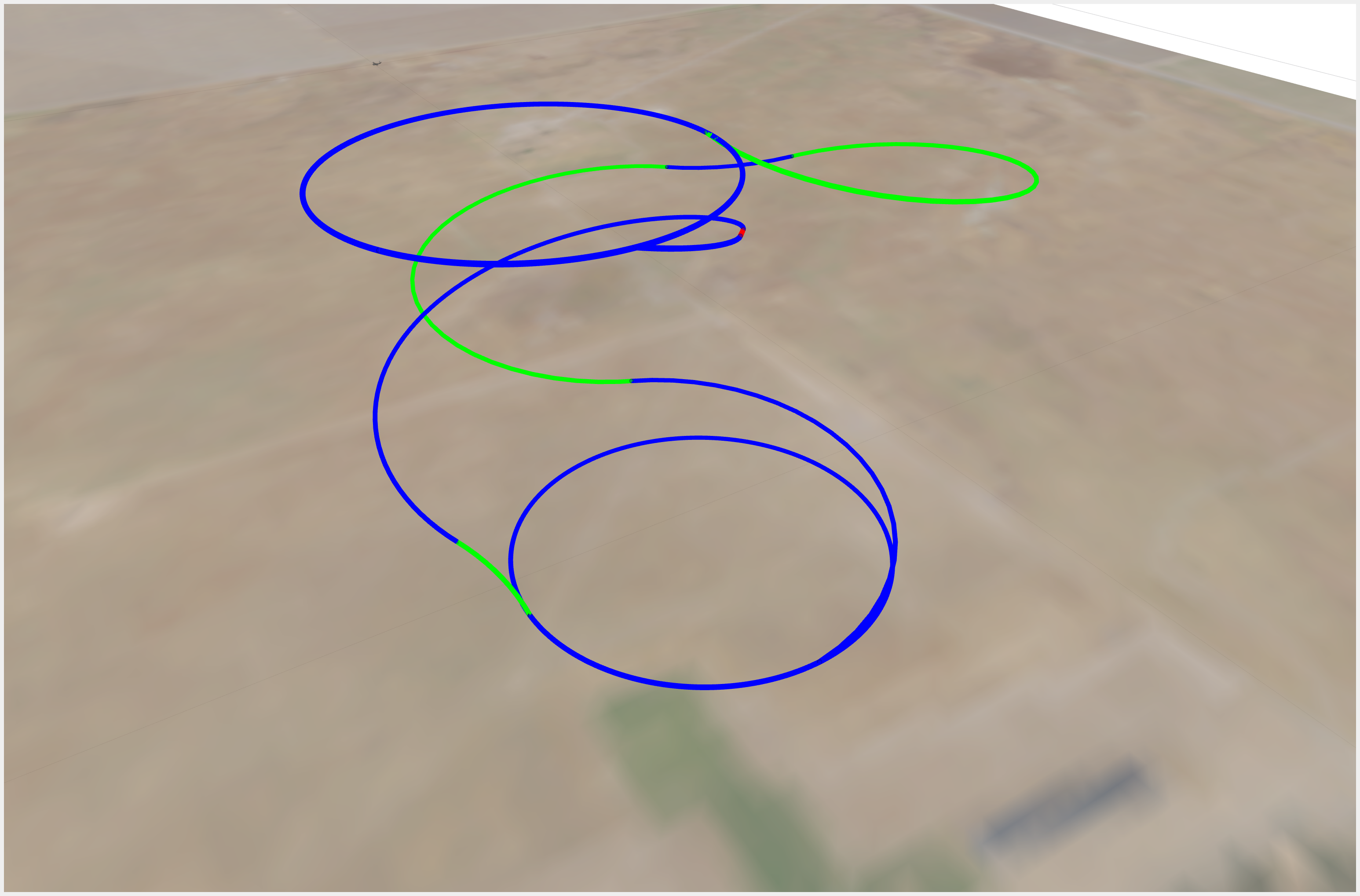}
    \caption{Reference Path}
    \label{fig:flight_data_reference_path}
\end{subfigure}
\begin{subfigure}{0.53\linewidth}
    \centering
    \includegraphics[width=\linewidth]{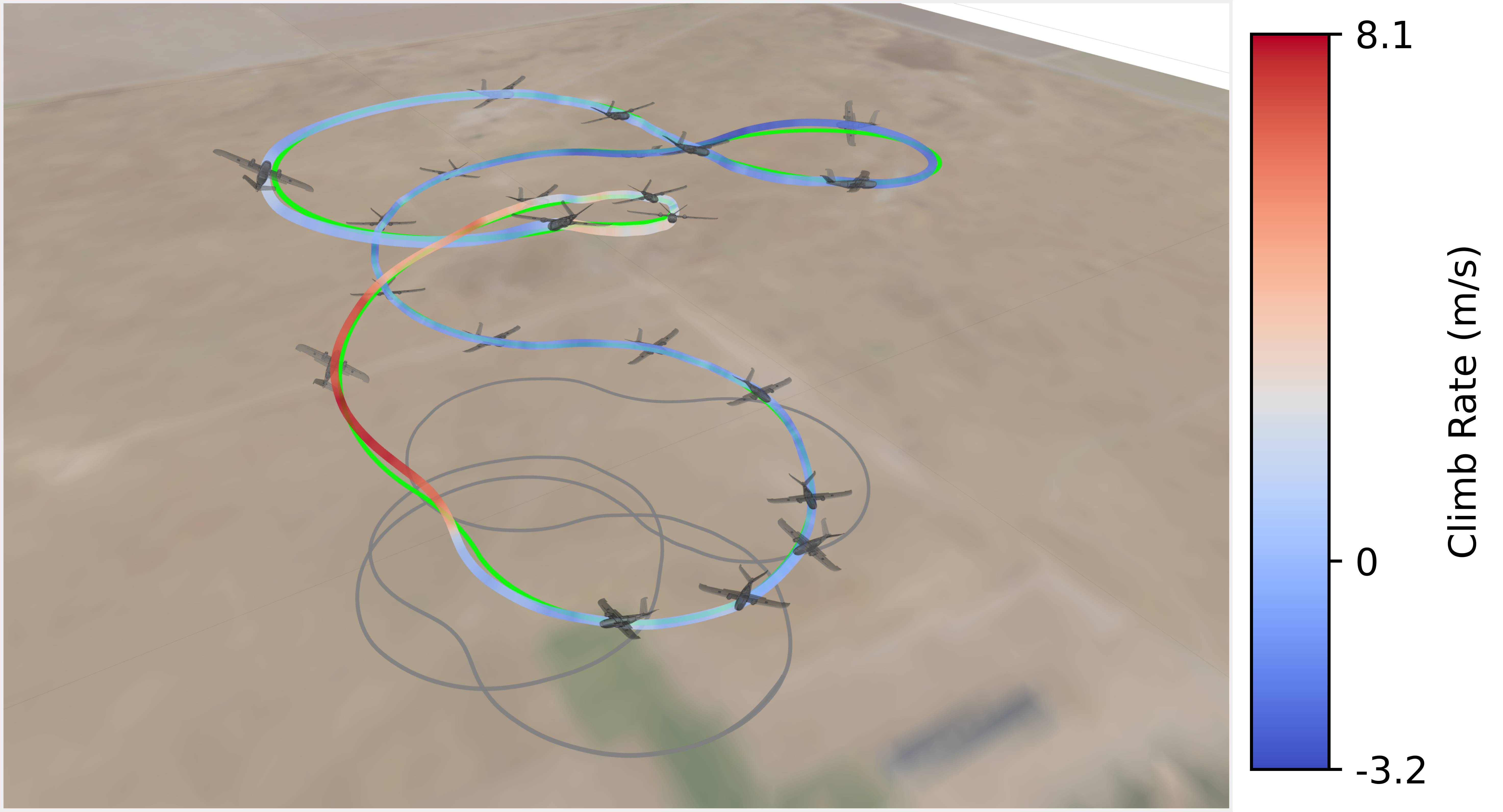}
    \caption{Vehicle Trajectory}
    \label{fig:flight_data_vehicle_trajectory}
\end{subfigure}
\caption{a) Reference Dubins airplane path colored with right turn (blue), left turn (green), straight segment (red). b) Vehicle trajectory following the path, colored with climb rate. The vehicle position and attitude is displayed every \SI{2}{\second}, and the reference path displayed as green.}
\label{fig:flight_data}
\end{figure*}

In order to evaluate the efficacy of the Dubins airplane paths, we consider the reduction in path length for randomly sampled initial and final states. We generate \qty{100,000}{} randomly generated samples, where the start position is fixed at the origin $\bm{x}_i=(0, 0, 0, \theta_i)$, and the orientation of the start state $\theta_i\in[-\pi, \pi)$, position and the orientation of the target state are sampled randomly $x_t\in[-4R_{min}, 4R_{min}]$, $y_t \in [-4R_{min}, 4R_{min}]$, $z_t [0, 4R_{min}]$, $\theta_t \in [-\pi, \pi)$. We consider two different vehicle models, \emph{Strix Stratosurfer} and \emph{FMX Fox}, where the minimum turn radius is both $R_{min}=\SI{66.67}{\metre}$, but differ at the climb rate limits as $u_{z_{max}}\in[-\SI{3}{\metre\per\second}, \SI{6}{\metre\per\second}]$, $u_{z_{max}}\in[-\SI{1.5}{\metre\per\second}, \SI{6}{\metre\per\second}]$ respectively. We compare the reduction of path length by dividing the minimum time path length of the asymmetric climb rate with the symmetric climb rate constraints. 

\reffig{fig:benchmark_path_length} shows the distribution of the ratio of path lengths of the two different models. The median of the path length ratio is \qty{1.72}{} for the \emph{Strix Stratosurfer}, and \qty{3.45}{} for the \emph{FMX Fox}. Note that the path length ratio is distributed mostly around two values. The first value is $1.0$, which represents the \emph{low altitude case} with no effect of the increased climb rate showing that accompanying larger climb rate values can never result in longer paths. The second value is the ratio of the increased climb rate~($2.0$, $4.0$), which corresponds to the \emph{medium altitude case} and \emph{high altitude case}. It can be seen that considering higher climb rate results in a significant reduction of path length, where it reduces by \qty{42}{\%} for the \emph{Strix Stratosurfer}, and by \qty{71}{\%} for the \emph{FMX Fox}. 
\subsection{Implications on Planning Performance}

We evaluate the implication for a downstream path planning problem of navigating through rugged terrain. The goal is to find a kinematically feasible path from the start and the goal, while staying within the distance limits~(\SI{50}{\metre} to \SI{120}{\metre}) to the terrain. A \acp{DEM} with the extent of $12.2~\si{km} \times 7.48~\si{km}$ with \SI{10}{\metre} lateral resolution is chosen, where it uses CH1903/LV03 coordinates, with Bessel 1841 as the vertical datum and provided by SwissAlti3D~\cite{swisstopo2023swissalti3d}. We fix the start and goal circle at $\bm{c}_{start} = (2992, -4720, \cdot)[m]$, $\bm{c}_{goal} = (-2992, 4880, \cdot)[m]$ relative to the center of the terrain map. The minimum turn radius of the vehicle is set at \SI{66.67}{\metre}, and the maximum climbrate as \SI{6}{\metre\per\second} and descend rate \SI{3}{\metre\per\second}, considering \SI{15}{\metre\per\second} ground speed. The Dubins airplane paths are used in an asymptoticly optimal sampling-based planner RRT*~\cite{karaman2010incremental}. 


Figure~\ref{fig:gotthard_pass} shows a qualitative comparison of the Dubins airplane paths between the start and the goal through a mountainous terrain after \SI{500}{\second} planning time. The path length with asymmetric climb rate (cyan) is \SI{14.37}{\kilo\metre}, while symmetric climb rate (red) is \SI{19.20}{\kilo\metre}. This demonstrate a significant reduction of solution path length with the added feasibility of the asymmetric climb rates. Note that the path more aggressively follows the terrain with the added feasibility of asymmetric climb rates from the start, but follows a similar path during descent as the climb rate constraints for descending is identical. 

Figure~\ref{fig:solution_convergence} shows that benchmarking the planner runs with 20 repeated runs with \SI{300}{\second} planning time results. The median time to find an initial solution is \SI{50.53}{\second} for the asymmetric Dubins path and \SI{139.90}{\second} for the symmetric Dubins path, showing that the time to find the initial solution is significantly reduced when using asymmetric Dubins paths. Moreover, it can be seen that the average solution path length found with asymmetric Dubins paths are significantly shorter. The benchmark was not run more than \SI{300}{\second} as it is practically not useful to plan onboard with longer solve times.

\section{Experimental Results}
In this section, we demonstrate the feasibility of deploying on a fixed-wing aerial vehicle.

\subsection{Setup}
\begin{figure}[t!]
    \vspace*{0.2cm}
    \centerline{\includegraphics[width=\linewidth]{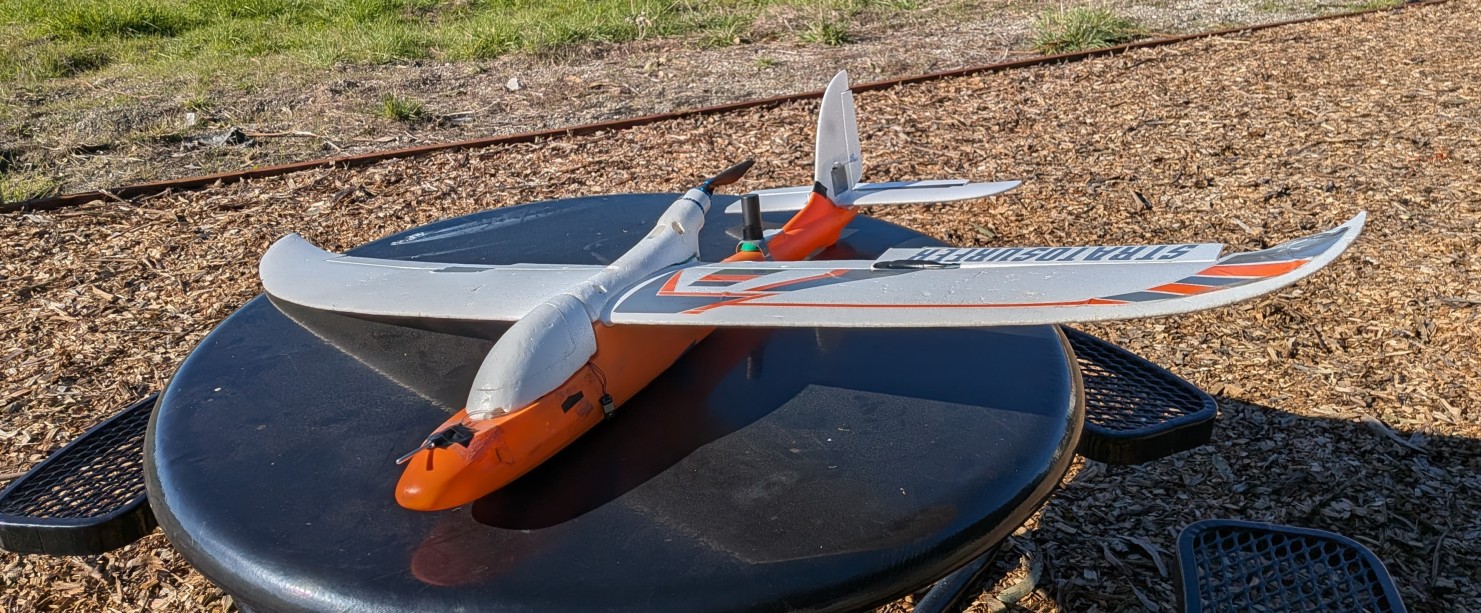}}
    \caption{Fixed-wing platform used for real world flight experiments.}
    \label{fig:test_vehicle}
        \vspace{-10pt}
\end{figure}
We use a fixed-wing platform based on a Strix Stratosurfer with a \SI{1.7}{\metre} wingspan and \SI{4}{\kilo\gram} weight~(\reffig{fig:test_vehicle}). The platform is equipped with \iac{FMU}, which handles low level control, and a mission computer which computes high level paths. The \ac{FMU} is a Holybro Pixhawk v6x, running PX4 Autopilot v1.16 which handles state estimation and path following control. The \ac{FMU} is connected via a \SI{433}{\mega\hertz} Sik Radio module to the ground control station, and an additional \SI{2.4}{\giga\hertz} remote control link to the safety pilot.
For the mission computer, NVIDIA Jetson Orin NX (Eight-core (ONX 16GB) Cortex A78AE Armv8.2(64-bit)) is used, where the mission computer is connected to the \ac{FMU} via UART. The mission computer runs Ubuntu 22.04 and the software is integrated into \ac{ROS} Humble. The \emph{Dubins airplane path} is computation is done with \ac{OMPL}~\cite{sucan_ompl_2012}. 

To generate the reference paths, two loiter paths with turn radius of \SI{80}{\metre} was placed with \SI{150}{\metre} altitude difference and \SI{75}{\metre} lateral spacing. The Dubins airplane path was generated from sampled states along each of the loiter paths, where the minimum turn radius is \SI{66.67}{\metre} and the maximum climb rate \SI{4.6}{\metre\per\second} and the descend rate \SI{2.26}{\metre\per\second} assuming \SI{15}{\metre\per\second} cruise speed. 
The path reference $\bm{r}=[\bm{p}, \bm{v}, \kappa]$, where $\bm{p}$ is the closest point, $\bm{t}$ is the tangent of the path, $\kappa$ is the curvature of the path, is sent to the guidance controller of the \ac{FMU} to follow the reference. The flight was performed at a flat field on a moderately windy day, averaging \SI{6.4}{\metre\per\second}.

\subsection{Results}
\begin{figure}[b]
    \centering
    \includegraphics[width=\linewidth]{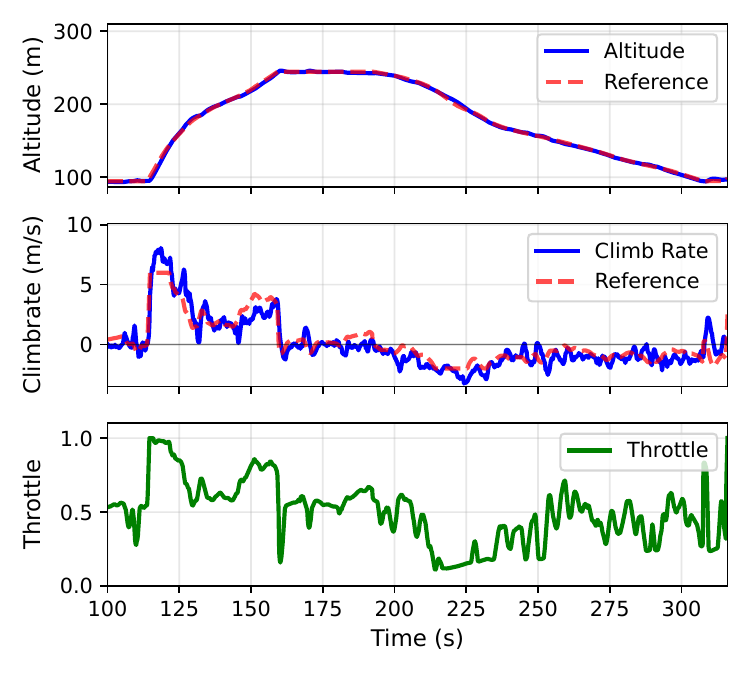}
    \caption{Flight data of the altitude, climbrate and throttle during the flight experiment.}
    \label{fig:flight_data_plots}
\end{figure}

Figure~\ref{fig:flight_data} shows the vehicle can successfully follow the reference path generated by Dubins airplane paths with asymmetric climb rates. The reference is the closest point of on the path from the vehicle. Longitudinally, it can be seen that the vehicle can follow the altitude reference precisely, resulting in \ac{RMSE} \SI{1.62}{\metre} of vertical and \SI{1.8}{\metre} lateral path tracking error (\reffig{fig:flight_data_plots}). It can be seen that the climb rate during descent is significantly lower than what can be achieved during climbing with high throttle, validating the asymmetry of the fixed-wing vehicle. Therefore, we can see that the vehicle can successfully follow the \emph{Dubins airplane paths} with asymmetric climb rates, and exploits full performance of the vehicle. Note that the throttle and climb rates vary from the expected rates due to the wind present during the flight, which ranged around \SI{5}{\metre\per\second}.

\section{Conclusion}
In this work, we propose \emph{asymmetric Dubins airplane paths}, which consider asymmetric climbrate constraints for the Dubins airplane paths. By exploiting the asymmetry of the climb rates, we show that we can significantly reduce the path duration by exploiting the full performance of fixed-wing vehicles. Moreover, it was shown that the added feasibility results in significantly improved solve time and shorter paths for a terrain planning task. The practicality of the approach is further demonstrated by a real world flight demonstration validating the asymmetric climb rates allow extracting full performance of the vehicle.

Future works will consider asymmetric Dubins airplane paths into real-time receding-horizon planning as well as consider more complex environmental effects such as wind. Finally, we aim to investigate compact path representation strategies by incorporating constraints and full vehicle dynamics.







\bibliographystyle{IEEEtran}
\bibliography{references.bib}

\end{document}